\def\year{2021}\relax
\documentclass[letterpaper]{article} 
\usepackage{aaai21}  
\usepackage{times}  
\usepackage{helvet} 
\usepackage{courier}  
\usepackage[hyphens]{url}  
\usepackage{graphicx} 
\usepackage{graphicx}  
\usepackage{natbib}  
\usepackage{caption} 
\usepackage{amsmath}
\usepackage{amssymb}
\usepackage{amsthm}
\usepackage{xcolor}

\usepackage{algorithm}
\usepackage{amsmath}
\usepackage{amssymb}
\usepackage[noend]{algpseudocode}
\usepackage{varwidth}
\usepackage{svg}
\usepackage{booktabs}
\usepackage{multirow}
\usepackage{siunitx}
\usepackage{subcaption}
\usepackage{comment}
\usepackage{tabularx}
\usepackage[switch]{lineno}
\usepackage{appendix}

\DeclareMathOperator*{\argmax}{arg\,max}

\newcommand{\mname}{{CUSTARD}} 
\newcommand*\patchAmsMathEnvironmentForLineno[1]{%
  \expandafter\let\csname old#1\expandafter\endcsname\csname #1\endcsname
  \expandafter\let\csname oldend#1\expandafter\endcsname\csname end#1\endcsname
  \renewenvironment{#1}%
     {\linenomath\csname old#1\endcsname}%
     {\csname oldend#1\endcsname\endlinenomath}}%
\newcommand*\patchBothAmsMathEnvironmentsForLineno[1]{%
  \patchAmsMathEnvironmentForLineno{#1}%
  \patchAmsMathEnvironmentForLineno{#1*}}%
\AtBeginDocument{%
\patchBothAmsMathEnvironmentsForLineno{equation}%
\patchBothAmsMathEnvironmentsForLineno{align}%
\patchBothAmsMathEnvironmentsForLineno{flalign}%
\patchBothAmsMathEnvironmentsForLineno{alignat}%
\patchBothAmsMathEnvironmentsForLineno{gather}%
\patchBothAmsMathEnvironmentsForLineno{multline}%
}

\title{Iterative Bounding MDPs: \\
Learning Interpretable Policies via Non-Interpretable Methods}
\author{
    Nicholay Topin,
    Stephanie Milani, 
    Fei Fang, 
    Manuela Veloso
}

\affiliations {
    \\Carnegie Mellon University\\
    Pittsburgh, PA 15213\\
    $\{$ntopin, smilani, feif, veloso$\}$@cs.cmu.edu
}
\begin{document}
\maketitle
\begin{abstract}
Current work in explainable reinforcement learning generally produces policies in the form of a decision tree over the state space. Such policies can be used for formal safety verification, agent behavior prediction, and manual inspection of important features. However, existing approaches fit a decision tree after training or use a custom learning procedure which is not compatible with new learning techniques, such as those which use neural networks. To address this limitation, we propose a novel Markov Decision Process (MDP) type for learning decision tree policies: Iterative Bounding MDPs (IBMDPs). An IBMDP is constructed around a base MDP so each IBMDP policy is guaranteed to correspond to a decision tree policy for the base MDP when using a method-agnostic masking procedure. Because of this decision tree equivalence, any function approximator can be used during training, including a neural network, while yielding a decision tree policy for the base MDP. We present the required masking procedure as well as a modified value update step which allows IBMDPs to be solved using existing algorithms. We apply this procedure to produce IBMDP variants of recent reinforcement learning methods. We empirically show the benefits of our approach by solving IBMDPs to produce decision tree policies for the base MDPs.
\end{abstract}

\section{Introduction}

The incorporation of deep neural networks into reinforcement learning (RL) has broadened the set of problems solvable with RL.
Though these techniques yield high-performing agents, the policies are encoded using thousands to millions of parameters, 
    and the parameters interact in complex, non-linear ways.
As a result, directly inspecting and verifying the resulting policies is difficult.
Without a mechanism for a human operator to readily inspect the resulting policy, we cannot deploy 
deep RL (DRL) 
in environments with strict regulatory or safety constraints.

Decision trees (DTs)
~\cite{quinlan1986induction} are an interpretable model family commonly used to represent policies.
Some benefits of DTs include that they allow formal verification of policy behavior~\cite{bastani2018viper}, counterfactual analysis~\cite{sokol2019desiderata}, and identification of relevant features.
However, DRL techniques are not directly compatible with policies expressed as DTs.
In contrast, some traditional RL algorithms, such as UTree~\cite{mccallum1997reinforcement}, 
produce DT policies but use specific internal representations that cannot be replaced with a deep neural network.
An alternative approach is to approximate a DRL policy with a DT, but the resulting policy can be arbitrarily worse than the original one
    and the DT can be large due to unnecessary intricacies in the original policy.
    \begin{figure}[t!]
         \begin{center}
         \includegraphics[width=0.99\columnwidth]{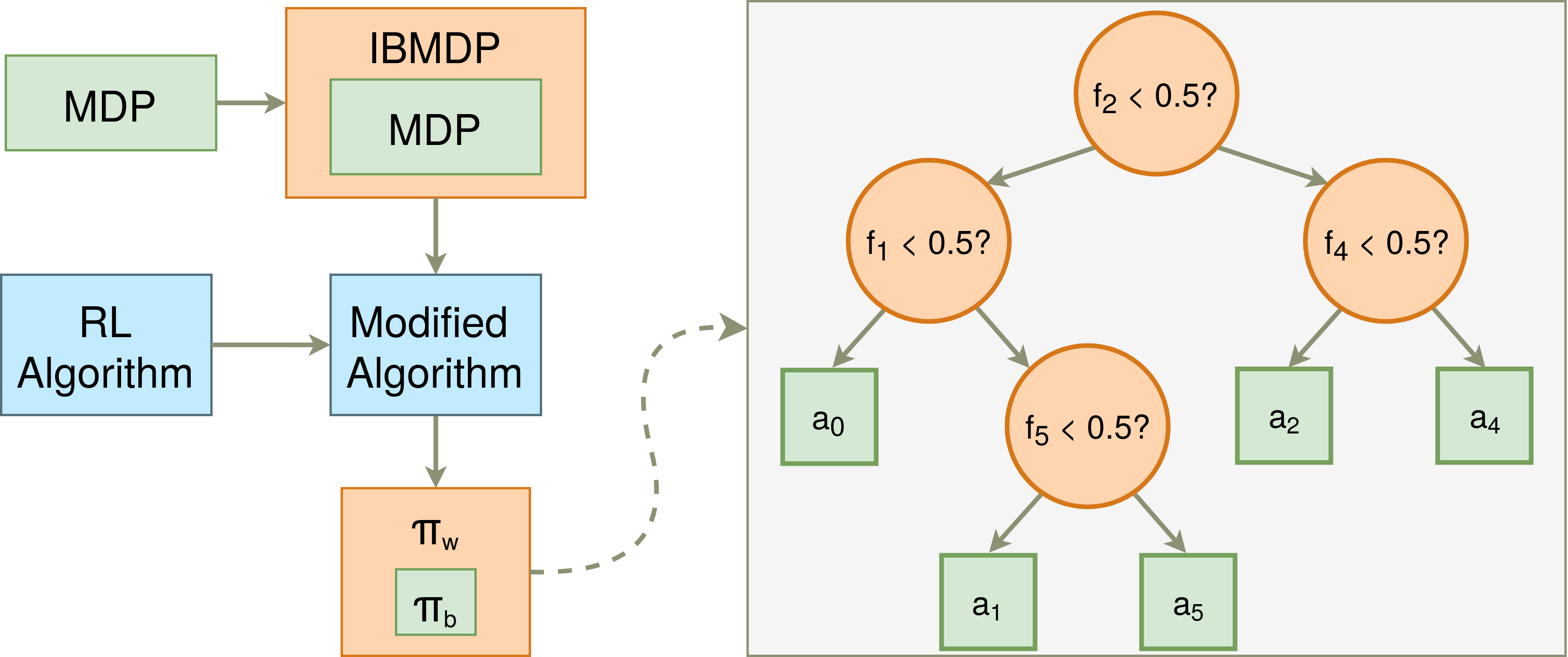}
         \end{center}
         \caption{Method overview: 
         we wrap a base MDP to form an IBMDP and 
         solve using a modified RL algorithm.
         The solution is a decision tree policy for the base environment.} 
         \label{fig:system}
    \end{figure}
    
To address the limitations of these techniques, we propose to solve a meta-problem using RL such that the solution corresponds to a DT-format policy for the original problem.
We introduce \mname{} (Constrain Underlying Solution to a Tree; Apply RL to Domain), a process that uses RL to solve the meta-problem while ensuring that the embedded solution is equivalent to a DT-format policy throughout training (overview in Figure~\ref{fig:system}).
We propose a novel Markov Decision Process (MDP) formulation for the meta-problem: Iterative Bounding MDPs.
We present a general procedure to make RL techniques compatible with \mname{}.
\mname{} allows modern DRL techniques to be applied to the meta-problem, 
    since the learned policy weights can be fully replaced by an equivalent DT.
Thus, \mname{} maintains the interpretability advantage of a DT policy while using a non-interpretable function approximator during training.
Additionally, \mname{} ensures that the DT is an exact representation of the learned behavior, 
    not an approximation.

The main contributions of this work are: 
(1) we introduce a novel MDP representation (IBMDPs) for learning a DT policy for a 
base MDP; 
(2) beginning with a two-agent UTree-like algorithm, we present an equivalent single-agent formulation which solves IBMDPs to produce DTs; 
(3) we show how to modify existing RL algorithms (policy gradient and Q-learning) to produce valid DTs for the base MDP; and 
(4) we empirically evaluate the performance of our approach and identify cases where it outperforms post-hoc DT fitting.

\section{Background}

\subsection{Solving Markov Decision Processes}
\label{sec:mdp}
In the RL framework, an agent acts in an environment defined by a Markov decision process (MDP).
An MDP is a five-tuple $\langle {S},{A},T,R, \gamma \rangle$, consisting of a set of states ${S}$, a set of actions ${A}$, a transition function $T$, a reward function $R$, and a discount factor $\gamma$. 
We focus on factored MDPs~\cite{boutilier1995exploiting}, in which each state consists of a set of feature value assignments $s = \{ f_1, ..., f_n \}$. Note that we do not require a factored reward function. 
An agent is tasked with finding a policy, $\pi : S \rightarrow A$, which yields the highest expected discounted future return for all states.
The expected return for a state $s$ when following policy $\pi$ is $V^\pi(s) = \mathbb{E}_\pi [\sum_{t=0}^\infty \gamma^t r_t]$.
Analogously, the expected return when taking action $a$ and following $\pi$ afterward is the Q-function $Q^\pi(s, a) = \mathbb{E}_\pi [r_0 + \gamma V^\pi(s')] = \mathbb{E}_\pi [r_0 + \gamma \max_{a'} Q^\pi(s',a')]$.

Q-learning-based algorithms directly approximate the Q-function, such as with a neural network~\cite{mnih2013playing}, and use it to infer the optimal policy $\pi^*$.
The Q-function estimate is incrementally updated to be closer to a \textit{target}, the bootstrapped estimate $r_t + \gamma \max_{a'} Q(s_{t+1},a')$.  
In contrast, policy gradient methods directly model and optimize the policy. 
Actor-critic methods~\cite{schulman2017proximal} additionally model the value function to leverage it in the policy update. 
They often use a critic for estimating the advantage function, $A(s,a) = Q(s,a) - V(s)$.

\subsection{Decision Trees in Reinforcement Learning}
Decision Trees (DTs) recursively split the input space along a specific feature based on a cutoff value, yielding axis-parallel partitions.
\textit{Leaf nodes} are the final partitions; \textit{internal nodes} are the intermediate partitions.
DT-like models have been used to represent the transition model~\cite{strehl2007efficient}, reward function~\cite{degris2006learning}, value function~\cite{pyeatt2001decision,tuyls2002reinforcement}, 
relative effect of actions~\cite{hester2010generalized}, and the policy~\cite{mccallum1997reinforcement}.
We focus on DT policies (DTPs), which map each state to a leaf node representing an action.

Sufficiently small DTPs are \textit{interpretable}~\cite{molnar2020interpretable}, in that people understand the mechanisms by which they work. 
DTs conditionally exhibit simulatability, decomposability, and algorithmic transparency~\cite{lipton2018mythos}.
When a person can contemplate an entire model at once, it is \textit{simulatable}; 
sufficiently small DTs exhibit this property.
A \textit{decomposable} model is one in which sub-parts can be intuitively explained; a DT with interpretable inputs exhibits this property.
\textit{Algorithmic transparency} requires an understanding of the algorithm itself: in particular, DTs are verifiable~\cite{bastani2018viper}, which is important in safety-critical applications.

\section{Related Work}
\subsection{Decision Tree Policies}
\label{sec:dtrl}
    Prior work on creating DTPs using arbitrary function approximators focuses on explaining an existing agent; 
    a non-DT policy is learned and then is approximated using a DT.
    One such method is VIPER~\cite{bastani2018viper}, which uses model compression techniques~\cite{bucilua2006model,hinton2015distilling,rusu2015policy} to distill a policy into a DT.
    This work adapts DAGGER~\cite{ross2011dagger} to prioritize gathering critical states, which are then used to learn a DT. 
    MOET~\cite{vasic2019moet} extends VIPER by learning a mixture of DTPs trained on different regions of the state space.
    However, both VIPER and MOET approximate an expert. When the expert is poorly approximated by a DT, the resulting DTPs perform poorly. 
    
    Other lines of work focus on directly learning a DTP, but they cannot use an arbitrary function approximator.
    UTree~\cite{mccallum1997reinforcement} and its extensions
    ~\cite{uther1998tree,pyeatt2003reinforcement,roth2019conservative}
    incrementally build a DT while training an RL agent.
    Transition tuples (or tuple statistics) are stored within leaf nodes, and a leaf node is split when the tuples suggest that two leaf nodes would better represent the Q-function.
    A concurrent work~\cite{rodriguez2020optimization} uses a differential decision tree to represent the policy and approximates the soft-max tree with a DT after training.
    However, these methods require specific policy or Q-function representations, so they cannot leverage powerful function approximators like neural networks.
    
    \subsection{Interpretable Reinforcement Learning}
    There exist other methods to produce interpretable RL policy summarizations. 
    One line of work produces graph-structured policies, including finite state machines~\cite{koul2018learning}, Markov chains~\cite{topin2019xdrl}, and transition graphs between 
    landmarks~\cite{sreedharan2020tldr}.
    Other work produces policies in custom representations.
    For example,~\citet{genetic} use a genetic algorithm to create policy function trees
    , which have algebraic functions as internal nodes and constants and state variables as leaves.
    \citet{HEIN2017fuzzyrl} express a policy as a fuzzy controller, which is a set of linguistic if-then rules whose outputs are combined.
    These policy formats address different aspects of interpretability compared to DTPs (e.g., showing long-term behavior rather than allowing policy verification).
    
    Another line of work uses attention mechanisms~\cite{wang2018reinforcement,annasamy2019towards,tang2020neuroevolution} to determine crucial factors in individual decisions made by the agent.
    A similar line of work is saliency methods, which produce visual representations of important pixels~\cite{greydanus2018visualizing,yang2018learn,huber2019enhancing} or objects~\cite{katia}.
    However,~\citet{atrey2019exploratory} argue that saliency maps are not sufficient explanations because the conclusions drawn from their outputs are highly subjective.
    Other methods explain decisions made by the agent as a function of the MDP components or the training process, including the reward function~\cite{anderson2019explaining,juozapaitis2019explainable,tabrez2019explanation}, 
    transition probabilities~\cite{cruz2019memorybasedxrl}, 
    and causal relationships in the environment~\cite{madumal2020distal,madumal2020explainable}. 
    These methods are orthogonal to our work; they provide different insights and can be used alongside our approach.
    
    \subsection{Hierarchical Reinforcement Learning}
    Our method can be viewed as a type of hierachical decomposition, similar to that performed in hierarchical RL~\cite{dayan1993feudal,dietterich2000maxq}.
    Perhaps the most well-known formulation is the options framework~\cite{precup1998theoretical,sutton1999between}, in which the problem is decomposed into a two-level hierarchy. 
    At the bottom level are options, which are subpolicies with termination conditions that take observations of the environment as input and output actions until the termination condition is met.
    A policy is defined over options; using this policy, an agent chooses an option and follows it until termination.
    Upon termination, the policy over options is again queried, and so on.
    Options over an MDP define a semi-MDP~\cite{bradtke1994reinforcement,mahadevan1997self,parr1998hierarchical}.
    In our method, the base MDP can be viewed as this semi-MDP and the IBMDP can be viewed as the full MDP.
    In a sense, the policy for the information-gathering actions is the lower-level policy, and the higher-level policy selects over the information-gathering policies.
    
\section{Approach}

    We present \mname{}, an approach for training an agent to produce a Decision Tree Policy using existing RL algorithms.
    We achieve this goal by training the agent to solve a wrapped version of the original, \textit{base} MDP. 
    The wrapped MDP, which we name an \textit{Iterative Bounding MDP}, extends the base MDP by adding information-gathering actions and \textit{bounding} state features to indicate the gathered information. 
    The bounding features correspond to a position within a DT traversal, and the information-gathering actions correspond to partitions performed by internal nodes within a DT.
    By constraining an agent's policy to be a function of the bounding state features, the learned policy is equivalent to a DT.
    
    In Section~\ref{approach_mdp_formulation}, we describe IBMDPs. 
    In Section~\ref{approach_tree_extraction}, we describe the process for extracting a DTP from an IBMDP policy during any point in training.
    In Section~\ref{approach_training_procedure}, we present methods for adapting existing RL algorithms to learn an implicit DTP for the base MDP. In particular, we describe modifications to Q-learning and actor-critic algorithms.

\subsection{Iterative Bounding MDPs} 
\label{approach_mdp_formulation}
    We introduce Iterative Bounding MDPs (IBMDPs), a novel MDP formulation for producing DTPs.
    We seek to produce a DTP by ensuring that an agent's IBMDP policy is equivalent to a DTP for the original, base MDP.
    To use a DTP to select an action in the base MDP, a series of internal nodes are traversed, and then the leaf node specifies the action.
    To allow this behavior, an IBMDP has actions that are equivalent to traversing nodes and state features that indicate the current node.
    
    The base MDP must be an MDP with a factored state representation, where each state feature has upper and lower bounds on its values.
    A \textit{base state} is a state from the base MDP's state space, and a \textit{wrapped state} is a state from the IBMDP's state space; other terms are defined analogously.
    
    \paragraph{State Space}
    A wrapped state $s_w$ consists of two parts: a base state $s_b$ and bounding features, $f_1^l$-$f_n^l$ and $f_1^h$-$f_n^h$.
    There exist two bounding features per base state feature, 
    such that $f_i^l$ represents a lower bound on the base feature $f_i$'s current value,
    and $f_i^h$ represents an upper-bound for that same base feature's current value. 
    The bounding features reflect the outcomes of binary comparisons performed during the traversal,
    and the bounds are tightened with more comparisons.
    A sequence of wrapped states represents a traversal through a DTP for a specific $s_b$. 
    For simplicity, and without loss of generality, we consider $s_b$ to be normalized such that all features are in $[0,1]$.
    We use $s_w[c]$ to refer to a component $c$ within $s_w$.
    An IBMDP state and state space are:
    \begin{equation*}
        S_w = S_b \times [0,1]^{2n}, ~~ s_w = \langle s_b, f_1^l, \dots, f_n^l, f_1^h, \dots, f_n^h \rangle.
    \end{equation*}
\paragraph{Action Space}
    The action space for an IBMDP $A_w$ consists of the base actions $A_b$ and an additional set of information-gathering actions $A_I$: 
    \begin{equation*}
        A_w = A_b \cup A_I.
    \end{equation*}
    Base actions correspond to taking an action within the base MDP, as when reaching a leaf in a DTP.
    Information-gathering actions specify a base state feature and a value, which correspond to the feature and value specified by an internal node of a DTP.
    We present two different action space formats: a discrete set of actions and a Parameterized Action Space~\cite{masson2016paramactionrl}.
    In both cases, the action can be described by a tuple, $\langle c, v \rangle$, where $c$ is the chosen feature and $v$ is the value.
    For simplicity, we consider $v \in [0,1]$, where $0$ and $1$ respectively correspond to the current lower and upper bound on $c$. 
    
    With a discrete set of IBMDP actions, each of the $n$ features can be compared to one of $p$ possible values.
    This results in $p \times n$ discrete actions, with $v$ values of $1/(p+1), \dots, p/(p+1)$ for each of the $n$ possible $f$.
    With this construction, the base actions must be discrete.
    In this case, the information-gathering actions are:
    
    \begin{equation*}
        A_I = \left\{ c_1, \dots, c_n \right\} \times \left\{ \frac{1}{p+1}, \dots, \frac{p}{p+1} \right\}.
    \end{equation*}
    In a Parameterized Action Space MDP (PASMDP), each action $a \in A_d$ has $m_a$ continuous parameters.
    A specific action choice is specified by selecting $(a, p_1^a, \dots, p_{m_a}^{a})$.
    If the IBMDP is a PASMDP, then there is an action for each of the $n$ features with a single parameter ($m_a=1$), where the action specifies $c$ and the parameter specifies $v$.
    With this formulation, the base MDP may have a continuous, multi-dimensional action space.
    This is supported by adding a single $a$ with parameters corresponding to the base action choices.
    If $A_b$ has discrete actions, then an $a$ is added for each of them, with the corresponding $m_a$ set to zero.
    The information-gathering actions in the PASMDP variant are:
    
    \begin{equation*}
        A_I = \left\{ c_1, \dots, c_n \right\} \times [0,1].
    \end{equation*}
\paragraph{Transition Function}
    When an agent takes an information-gathering action, $\langle c, v \rangle$, the selected value $v$ is compared to the indicated feature $c$.
    Since $v$ is constrained to $[0,1]$ but represents values in $[c^l, c^h]$, the un-normalized $v_p$ is obtained by projecting $v_p \leftarrow v \times (c^h - c^l) + c^l$.
    The bounding features $c^l$ and $c^h$ are updated to reflect the new upper- and lower-bounds for $c$;
    the base features are unchanged.
    This process is equivalent to the behavior of an internal node in a DTP: a feature is compared to a value, and the two child nodes represent different value ranges for that feature.
    Thus, for an information-gathering action $\langle c, v \rangle$, the transition function of the IBMDP, $T_w$, is deterministic, and the next state, $s_w'$, is based on $s_w$:
    \begin{equation*}
    \begin{aligned}
    &s_w'[s_b] = s_w[s_b], \\
    &s_w'[f] = s_w[f] \forall f \notin \{c^l, c^h\}, \\
    &\textrm{If } s_b[c] \leq v_p \textrm{: } s_w'[c^h] = \min(s_w[c^h], v_p), s_w'[c^l] = s_w[c^l], \\
    &\textrm{If } s_b[c] > v_p \textrm{: } s_w'[c^l] = \max(s_w[c^l], v_p), s_w'[c^h] = s_w[c^h].
    \end{aligned}
    \end{equation*}
    When a base action is taken, the base features are updated as though this action was taken in the base MDP, and the bounding features are reset to their extreme values. This is equivalent to selecting a base action in a DTP and beginning to traverse the DTP for the next base state (starting from the root node). This corresponds to a transition function of:
    \begin{equation*}
    \begin{aligned}
    &a \in A_b \land ((s_w'[f_i^l] = 0) \land (s_w'[f_i^h] = 1) \forall i \in \left\{1, \dots, n\right\} ) \\
    &\Longrightarrow T_w(s_w, a, s_w') = T_b(s_w[s_b], a, s_w'[s_b]).
    \end{aligned}
    \end{equation*}
\paragraph{Reward Function}
    The reward for a base action is the reward specified by the base MDP for the base action, base original state, and base new state. 
    The reward for information-gathering actions is a fixed, small penalty $\zeta$.
    For a sufficiently low value of $\zeta$, the optimal solution for the IBMDP includes the optimal solution of the base MDP.
    The overall IBMDP reward function is:
    \begin{equation*}
    \begin{aligned}
    a \in A_b &\implies R(s_w, a, s_w') = R(s_w[s_b], a, s_w'[s_b']),  \\
    a \notin A_b &\implies R(s_w, \langle c, v \rangle, s_w') = \zeta. 
    \end{aligned}
    \end{equation*}
    \paragraph{Gamma}
    We introduce a second discount factor, $\gamma_w$.
    When a base action is taken in the IBMDP, the gamma from the base MDP, $\gamma_b$, is used to compute the expected discounted future reward.
    Otherwise, $\gamma_w$ is used.
    For a $\gamma_w$ sufficiently close to $1$, the expected discounted future reward is identical for an $s_w$, if acted upon in the IBMDP, and its corresponding $s_b$, if acted upon in the base MDP.
\paragraph{Remaining Components}
    We present the additional components required for an episodic MDP, but the framework is also applicable to non-episodic environments. 
    A transition in the IBMDP, $(s_w, a_w, s_w')$, is terminal if $a \in A_b$ and $(s_w[s_b], a, s_w'[s_b])$ is a terminal transition in the base MDP.
    The distribution over starting states of the IBMDP is derived from the distribution of starting states in the base MDP. 
    The probability of starting in state $s_w$ is $0$ if some $f_i^l \neq 0$ or $f_i^h \neq 1$; otherwise, it is equal to the probability of starting in $s_w[s_b]$ in the base MDP.

\subsection{Tree Extraction} \label{approach_tree_extraction}
    \begin{algorithm}[t] 
    \caption{Extract a Decision Tree Policy from an IBMDP policy $\pi$, beginning traversal from $obs$.} \label{alg_extract_tree}
    \begin{algorithmic}[1]
      \Procedure{Subtree\_From\_Policy}{$obs, \pi$} 
        \State $a \leftarrow \pi(obs)$ \label{pseudo:act}
        \If{ $a \in A_b$} \Comment{Leaf if base action} \label{pseudo:act2}
            \State $\textbf{return } \textrm{Leaf\_Node}(\textrm{action: } a)$ \label{pseudo:leaf}
        \Else
            \State $c,v \leftarrow a$ \Comment{Splitting action is feature and value} \label{pseudo:split}
            \State $v_p \leftarrow v \times (obs[c^h] - obs[c^l]) + obs[c^l]$ \label{pseudo:project}
            \State $obs_L \leftarrow obs; \qquad obs_R \leftarrow obs$ \label{pseudo:child}
            \State $obs_L[c^h] \leftarrow v_p; \quad obs_R[c^l] \leftarrow v_p$ \label{pseudo:child2}
            \State $child_L \leftarrow \textrm{Subtree\_From\_Policy}(obs_L, \pi)$ \label{pseudo:recurse}
            \State $child_R \leftarrow \textrm{Subtree\_From\_Policy}(obs_R, \pi)$ \label{pseudo:recurse2}
            \State $\textbf{return } \textrm{Internal\_Node}(\textrm{feature: } c, \textrm{value: } v_p,$ \par
            \hspace{4.5em}  $\textrm{children: } (child_L, child_R) )$ \label{pseudo:final}
        \EndIf
      \EndProcedure
    \end{algorithmic}
    \end{algorithm}

    Not all policies for the IBMDP correspond to valid DTPs; the presence of $s_b$ within each wrapped state allows access to full state information at any point during tree traversal.
    However, all IBMDP policies that only consider the bounding features (i.e., ignore $s_b$) correspond to a DTP.
    We describe the process for extracting a DTP from a policy defined over bounding observations from the environment, $\pi(s_w \setminus s_b)$. We present the training of such policies in Section~\ref{approach_training_procedure}.
    
    Algorithm~\ref{alg_extract_tree} outlines the full DTP extraction procedure.
    $\textsc{Subtree\_From\_Policy}$ constructs a single node based on the current observation; that node's children are constructed through recursive calls to this same function.
    As described in Section~\ref{approach_mdp_formulation}, the bounding features ($s_w \setminus s_b$) describe a node within a DTP, with $s_w[f_i^l] = 0 \land s_w[f_i^h] = 1 \forall i \in [1, \dots, n]$ corresponding to the root node.
    $\textsc{Subtree\_From\_Policy}(s_w \setminus s_b, \pi)$ for a root node $s_w$ yields the DTP for $\pi$.
    
    An action $a$ within the IBMDP corresponds to a leaf node action (when $a \in A_b$) or a DT split (when $a \notin A_b$).
    Lines~\ref{pseudo:act}-\ref{pseudo:act2} obtain the action for the current node and identify its type.
    The action taken for a leaf node defines that leaf, so Line~\ref{pseudo:leaf} constructs a leaf if $a$ is not an information gathering action.
    Information gathering actions consist of a feature choice $c$ and a splitting value $v$ (Line~\ref{pseudo:split}).
    The IBMDP constrains $v$ to be in $[0,1]$, which corresponds to decision node splitting values between $s_w[c^l]$ and $s_w[c^h]$, the current known upper and lower bounds for feature $c$.
    Line~\ref{pseudo:project} projects $v$ onto this range, yielding $v_p$, to which feature $c$ can be directly compared.
    
    To create the full tree, both child nodes must be explored, so the procedure considers both possibilities ($s_b[c] \leq v_p$ and $s_b[c] > v_p$).
    Lines~\ref{pseudo:child}-\ref{pseudo:child2} construct both possible outcomes: a tighter upper bound, $c^h \leftarrow v_p$, and a tighter lower bound, $c^l \leftarrow v_p$.
    This procedure then recursively creates the child nodes (Lines~\ref{pseudo:recurse}-\ref{pseudo:recurse2}).
    The final result (Line~\ref{pseudo:final}) is an internal DTP node: an incoming observation's feature is compared to a value $v_p$ ($obs[c] \leq v_p$), and traversal continues to one of the children, depending on the outcome of the comparison.

\subsection{Training Procedure} \label{approach_training_procedure}
    If an agent solves an IBMDP without further constraints, then it can learn a policy where actions depend on $s_b$ in arbitrarily complicated ways.
    To ensure that the base MDP policy follows a DT structure, the IBMDP policy must be a function of only the bounding features.
    Effectively, if the policy is a function of $s_w \setminus s_b$, then the policy is a DTP for the base MDP.
    However, with a policy of the form $\pi(s_w \setminus s_b)$, the standard bootstrap estimate does not reflect expected future reward
    because the next observation is always the zero-information root node state.
    Therefore, standard RL algorithms must be modified to produce DTPs within an IBMDP.
    
    We present a set of modifications that can be applied to standard RL algorithms so the one-step bootstrap reflects a correct future reward estimate. 
    We motivate this set of modifications by presenting a ``two agent'' division of the problem and then show the equivalent single-agent $Q$ target.
    We then demonstrate how a target Q-function or critic can be provided the full state ($s_w$) to facilitate learning while maintaining a DT-style policy.
    Finally, we present how the modifications are applied to Q-learning and actor-critic algorithms.
    Without loss of generality, we focus on learning a Q-function.
    If learning an advantage function or value function, 
    an analogous target modification can be made.
    
    \paragraph{Two Agent Division}
    \begin{figure}[t!]
         \begin{center}
         \includegraphics[width=0.95\columnwidth]{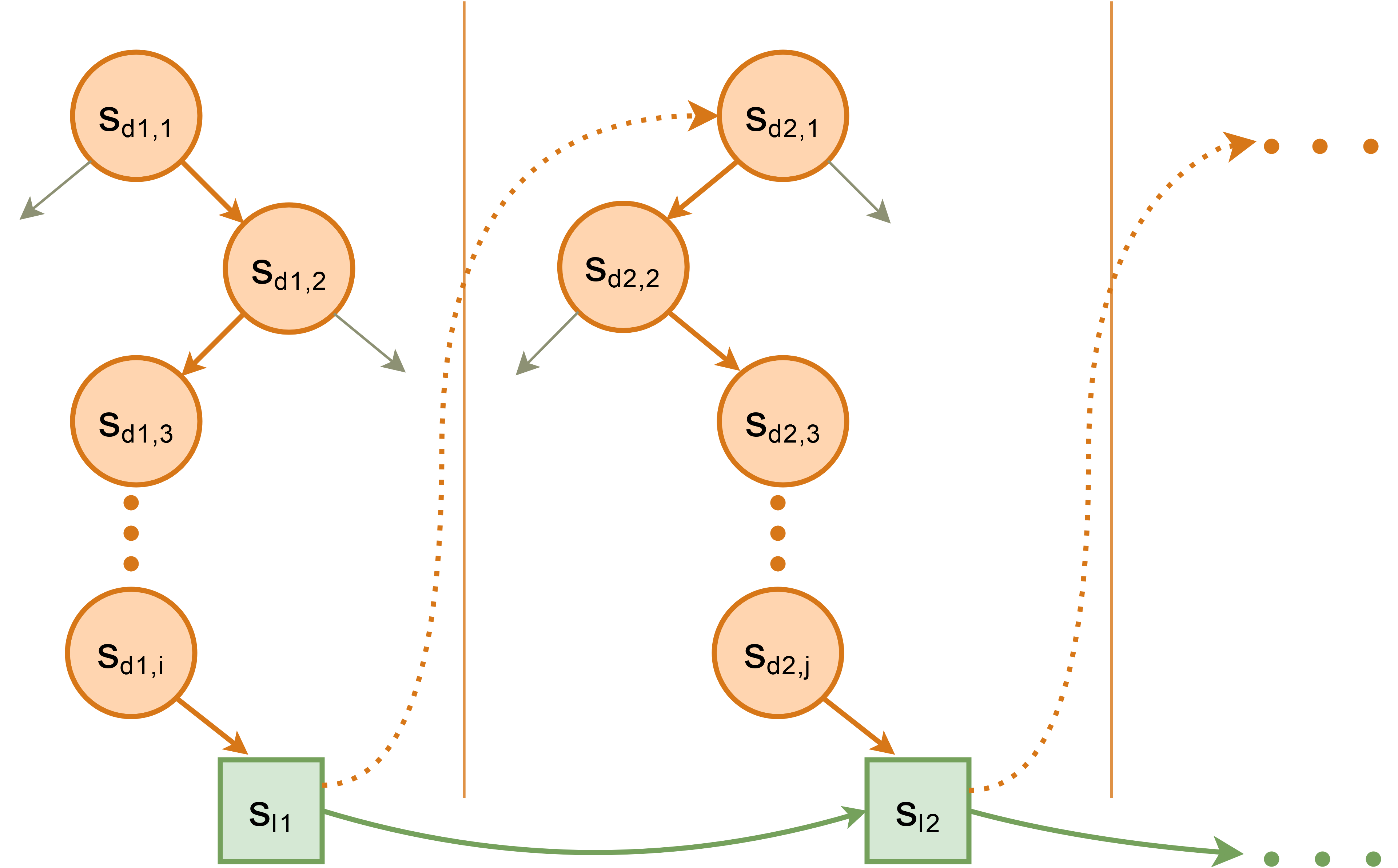}
         \end{center}
         \caption{The division between the tree agent (orange circle states and arrow actions) and the leaf agent (green square states and arrow actions). Each tree traversal is an episode for the tree agent and one transition for the leaf agent.} 
         \label{fig:2agent}
    \end{figure}
    Learning in an IBMDP can be cast as a two-agent problem:
    (i) a \textit{tree agent} selects which information-gathering actions to take and when to take a base action, and
    (ii) a \textit{leaf agent} selects a base action using the bounding features, when prompted to do so.
    Figure~\ref{fig:2agent} shows this division, where the leaf agent selects actions in $s_{l_1}$ and $s_{l_2}$, and the tree agent selects all other actions.
    
    With this division of the problem, the leaf agent is equivalent to the agent in UTree-style methods. 
    The tree agent replaces the incremental tree construction used in UTree and is akin to an RL agent constructing a DT for a supervised problem~\cite{preda2007buildingdtbyrl}.
    The leaf agent's observed transition sequence consists of leaf nodes and its own selected actions: $s_{l_1}, a_{l_1}, r_{l_1}, s_{l_2}, a_{l_2}$.
    The bootstrapped Q-value estimate is: 
    \begin{equation*}
    r_{l_1} + \gamma_b \max_{a' \in A_b} Q_l(s_{l_2}, a'),
    \end{equation*}
    where $r_{l_1}$ is a reward obtained from the base MDP.
    
    In this framing, the tree agent experiences a new episode when a base action is taken.
    The initial state is always the zero-information, root state, and the episode terminates when the agent chooses the stop splitting action, $a_{stop}$, which we add for the two agent formulation.
    When the tree agent stops splitting, the reward is the value estimated by the leaf agent, $Q_l(s_l, a_l)$.
    The tree agent's Q-value target is:
    \begin{equation*}
    r_{d} + \gamma_w \max_{a' \in a_{stop} \cup A_w \setminus A_b} Q_d(s_{d}', a'),
    \end{equation*}
    where $r_{d}$ is $\max_{a' \in A_b} Q_l(s_{d}, a')$ if the $a_{stop}$ action was chosen and $\zeta$ otherwise.
    When $a_{stop}$ is taken, $Q_d(s_{d}', a')$ is $0$ for all $a'$ since the transition is terminal for the tree agent.
    
    These two equations for target Q-values allow an IBMDP to be solved using only the partial $s_w \setminus s_b$ observations.
    The tree agent does not directly receive a reward signal from future base actions but uses the leaf agent's estimates to update. The leaf agent learns long-term reward estimates based on rewards from the environment.
    
    \paragraph{Merging of Agents}
    The target Q-value for a terminal tree agent action is $r_{d}$, which is $\max_{a \in A_b} Q_l(s, a)$. 
    The tree agent's episode terminates if and only if $a_{stop}$ is taken.
    Effectively, the tree agent seeks to learn $Q_d(s, a_{stop}) = \max_{a \in A_b} Q_l(s, a)$.
    Rather than learning this relationship, $Q_d(s, a_{stop})$ can directly query $Q_l$, simplifying the learning task without changing the underlying problem.
    
    With this change to $Q_d(s, a_{stop})$, $Q_d$ and $Q_l$ are defined over disjoint subsets of $A_w$.
    A single, unified Q-function $Q$ can be learned, which is defined over all $a$ in $A_w$.
    This allows the target Q-values to be re-written as:
    
    \begin{equation*}
    \begin{aligned}
    a \in A_b \implies \textrm{target } &= r_{l_1} + \gamma_b \max_{a' \in A_b} Q(s_{l_2}, a'), \\
    a \notin A_b \implies \textrm{target } &= \zeta + \gamma_w \max_{a' \in A_w} Q(s', a'),
    \end{aligned}
    \end{equation*}
    where $s'$ is the next state, regardless of type.
    In the former equation, $s_{l_2}$ is the next state in which a base action is taken when following the greedy policy.
    In the latter equation, if the $\max$ returns the Q-value for an $a \notin A_b$, the two terms correspond to the reward and expected discounted future reward.
    When the $\max$ returns the Q-value for an $a \in A_b$, the two terms are then the immediate reward and the reward from $a_{stop}$ in the next state,
    effectively removing the terminal/non-terminal distinction for the tree agent.
    
    As a result, this two-agent problem is equivalent to a single agent updating a single Q-function using two different targets, depending on the action taken.
    The equation for computing a target differs from the standard Q-function update equation (as applied to the IBMDP) in one way:
    if a base action is taken, the ``next state'' is the next state in which a base action is taken, rather than simply the next state.
    This single change is sufficient to learn DTPs for IBMDPs.
    
    \paragraph{Omniscient Q-function}

    The above merged agent formulation can be directly used to learn DTPs.
    However, the merged formulation requires the next leaf state, $s_{l_2}$, when a base action is taken.
	This state is not naturally encountered when performing off-policy exploration, so $s_{l_2}$ must be computed by repeatedly querying the Q-function with a sequence of $s_d$ tree states until the next base action is chosen.
    As a result, computing a single base action target Q-value requires simulating the next choice of base action, roughly doubling the computation time.
	
	As an extention of the merged agent formulation, we propose to approximate $Q(s_{l_2}, a)$ using a second Q-function, $Q_o$.
	We refer to this second Q-function as an \textit{omniscient Q-function}, because its input is the full state $s_w$.
	$Q_o$ is used in a supporting role during training; the policy is obtained directly from $Q$. 
	As a result, providing $Q_o$ the full state, $s_w$, does not violate the extraction process's requirement that the policy is a function of only $s_w\setminus s_b$. 
    
	The omniscient Q-function is trained to approximate $Q(s_{l_2}, a)$ based on $a$ and the full state at $s_{l_2}$'s root, $s_r$. 
	This root state is sufficient since $s_{l_2}$ is obtained from $s_r$ through a sequence of actions, each based on the previous $s_d$. 
	Therefore, the current greedy policy corresponds to some function $F(s_r) = s_{l_2}$ for all $(s_r, s_{l_2})$ pairs.
	We have $Q_o$ implicitly learn this function as it aims to learn an approximation $Q_o(s_r, a) \approx Q(s_{l_2}, a)$ for all base actions.
    
    Additionally, the original merged formulation learns the Q-value at each level in the tree (for $s_{d1,1}, s_{d1,2},$ etc.) using targets computed from the next level.
    This leads to slow propagation of environment reward signals from the leaf nodes.
	In addition to using $Q_o$ for the root node, we propose to have it learn to approximate $Q_o(s_w, a) \approx Q(s, a)$ for all states and all actions.
    Since $Q_o$ has access to $s_b$, the rewards obtained in the leaf node, $s_{l_1}$, directly propagate through $Q_o$ to earlier levels of the tree instead of sequentially propagating upward (from leaf to root).
	\begin{figure}[t]
    \centering
    \includegraphics[width=\columnwidth]{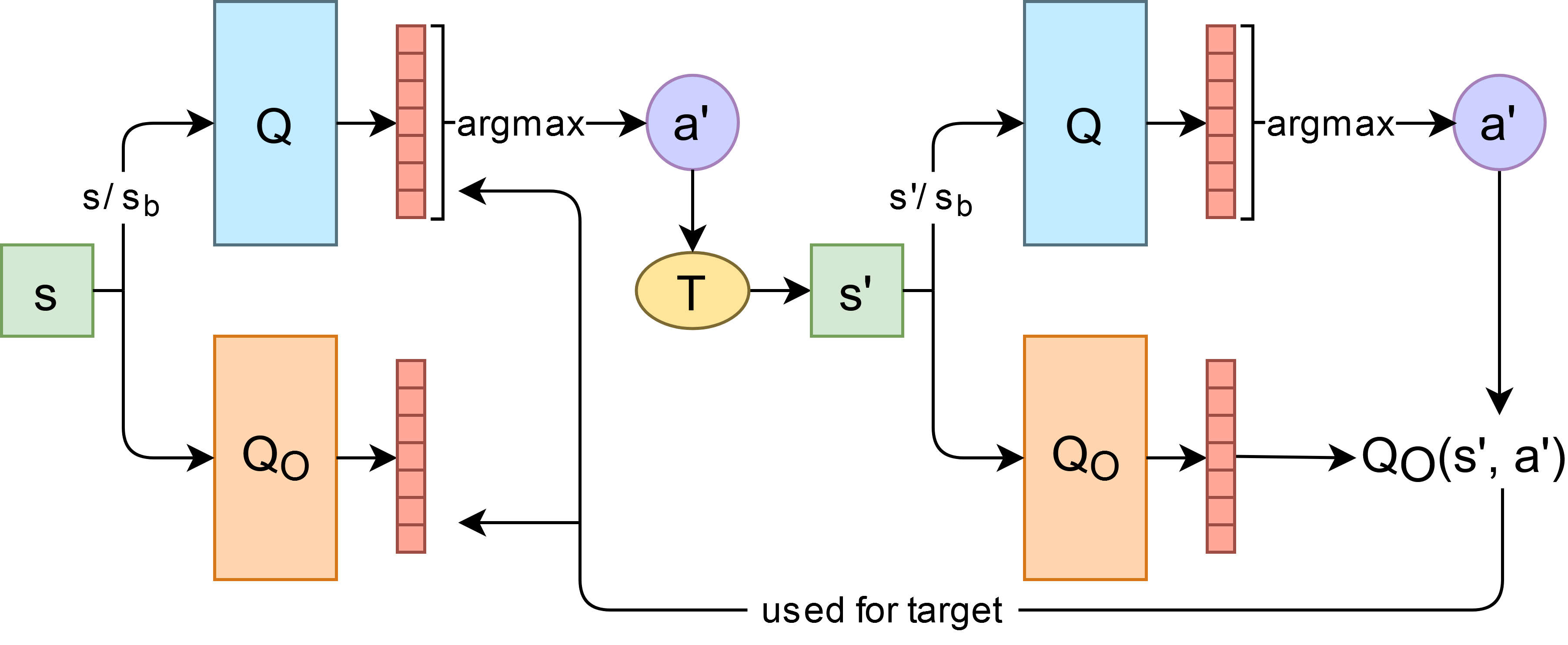}
    \caption{The method for using the omniscient Q-function, $Q_o$, for $Q$ targets. The policy is based only on $Q$, so a DTP can be extracted despite $Q_o$ being a function on the full state.}
    \label{fig:omnqfn}
\end{figure}
	
    As shown in Figure~\ref{fig:omnqfn}, during training, we use $Q_o$ in cases where $Q(s, a)$ would be used as a target.
    The action choice is still based on $Q(s,a)$, but the value is obtained from $Q_o$.
    Both $Q_o$ and $Q$ are updated using the $Q_o$-based target.
    
    \paragraph{Modifying Standard RL Algorithms}
    For Q-learning-based methods, such as Dueling Deep Q-Networks (DDQN)~\cite{wang2016dueling} and Model-Free Episodic Control (MFEC)~\cite{blundell2016model},
    we use the merged agent formulation target value of $Q_o(s_w, \argmax_a Q(s_w \setminus s_b, a))$ in place of $\max_a Q(s,a)$ (for both $a \in A_b$ and $a \notin A_b$);
    the additional Q-function, $Q_o$, is updated using the same target value.
    For policy gradient methods, such as Proximal Policy Optimization (PPO)~\cite{schulman2017proximal}, the Q-function is used to compute advantage values only during training.
    Therefore, we use only $Q_o$, not $Q$, to compute advantages.
    $Q_o$ is then trained using the merged agent formulation target value computations (replacing $Q(s,a)$ with $Q_o(s_w, a)$).

\section{Experiments}
We evaluate \mname{}'s ability to generate DTPs through solving an IBMDP using a non-interpretable function approximator during the learning process.
An alternative to implicitly learning a DTP is to learn a non-tree expert policy and then find a tree that mimics the expert.
We compare to VIPER, which takes this alternative approach and outperforms standard imitation learning methods. 
VIPER gathers samples using a DAGGER variant and weights the samples during tree training.
For this evaluation, we use three environments, briefly described in Section~\ref{sec:envs}, with further details in the Appendix.

\subsection{Environments}
\label{sec:envs}

\paragraph{CartPole} 
    CartPole~\cite{brockman2016gym} is a commonly used domain for evaluating methods that produce DTPs, where the agent must balance a pole affixed to a cart by moving the cart back and forth.  
    An episode terminates when the pole falls or $200$ timesteps have elapsed.
    We include it to provide standard benchmark results; following previous work, methods are limited to DTs of depth two.

\paragraph{PrereqWorld}
    PrereqWorld~\cite{topin2019xdrl} is an abstraction of a production task; the agent is tasked with constructing a goal item.
    Creating each item may require a subset of other, prerequisite items.
    This environment is similar in structure to advising domains~\cite{dodson2011natural} and crafting, as in MineRL~\cite{milani2020minerl}.
    The items are topologically sorted based on their prerequisite relationships such that lower-numbered items require higher-numbered items.
    A PrereqWorld environment has a number of item types $m$.
    A state consists of $m$ binary features which indicate whether each type of item is present.
    There are $m$ actions: each action corresponds to an attempt to produce a specific item.
    Creating the goal item yields a reward of $0$ and ends the episode; other actions yield a reward of -1.
    We use a base PrereqWorld environment with $m=10$ with a fixed set of prerequisites (specified in the Appendix).
    We produce smaller variants by removing high-numbed items (based on the topological sort).

\paragraph{PotholeWorld}
    We introduce the PotholeWorld domain, where
    the agent is tasked with traversing $50$ units along a three-lane road.
    The first lane gives less reward per unit traveled, but the other two lanes contain ``potholes'' which lead to a reward penalty if traversed.
    A state contains a single feature: the current position, in $[0,50]$. The initial state is at position $0$; an episode terminates when the position is $50$.
    The three actions are $[lane\_1, lane\_2, lane\_3]$, which each advance the agent a random amount drawn from $Unif(0.5,1)$.
    Potholes are added starting from position $0$ until position $50$ with distances drawn from $Unif(1,2)$. Potholes are assigned with equal probability to lane two or lane three. 
    When an action is taken, the base reward is equal to the distance moved. This is reduced by $10\%$ if the $lane\_1$ action was taken.
    If a pothole is in the chosen lane along the traversed stretch (i.e., between previous agent position and next agent position),
    then the reward is reduced by $5$.
    We specify the pothole positions used in our evaluation in the Appendix.
    
\begin{table*}[t]
\centering 
  \begin{tabular}{p{3.2cm}p{2.4cm}p{2.4cm}p{2.4cm}p{2.4cm}p{2.4cm}p{2.4cm}}
    \toprule
    \multirow{2}{1cm}{} &
      \multicolumn{1}{l}{CartPole} &
      \multicolumn{2}{l}{PrereqWorld} &
      \multicolumn{2}{l}{PotholeWorld} \\
      
      & {Reward} & {Reward} & {Depth} & {Reward} & {Depth} \\
      \midrule
    VIPER (DQN) & 200.00~~(0.00) & -4.00~~(0.00) & 5.70~~(0.62) & 46.30~~(0.39) & 6.00~~(0.61)\\
    VIPER (BI) & 200.00~~(0.00) & -4.00~~(0.00) & 6.00~~(0.00) & 46.31~~(1.32) & 9.18~~(1.08) \\
    \mname{} (DQN) & 198.72~~(4.74) & -4.08~~(0.34) & 4.28~~(0.67) & 46.92~~(2.14) & 5.36~~(1.41) \\
    \mname{} (PPO) & 199.32~~(3.23) & -4.04~~(0.20) & 4.16~~(0.47) & 45.39~~(0.42) & 1.04~~(0.75) \\
    \mname{} (MFEC) & 200.00~~(0.00) & -4.00~~(0.00) & 3.92~~(0.27) & 49.18~~(1.04) & 9.74~~(0.49) \\
    \bottomrule
  \end{tabular}
\caption{Final average reward and tree depth (Std Dev).}
\label{table:mname_table}
\end{table*}

\normalsize
\subsection{Learning with \mname{}}
\label{sec:mname_learning}
\begin{figure}[t]
    \centering
    \begin{subfigure}[t]{0.95\columnwidth}
         \centering
         \includegraphics[width=\columnwidth]{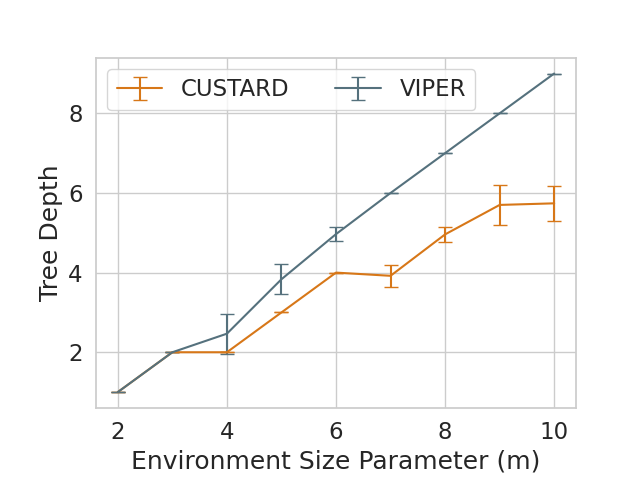}
         \label{fig:sizevsdepth}
     \end{subfigure}
     \begin{subfigure}[t]{0.95\columnwidth}
         \centering
         \includegraphics[width=\columnwidth]{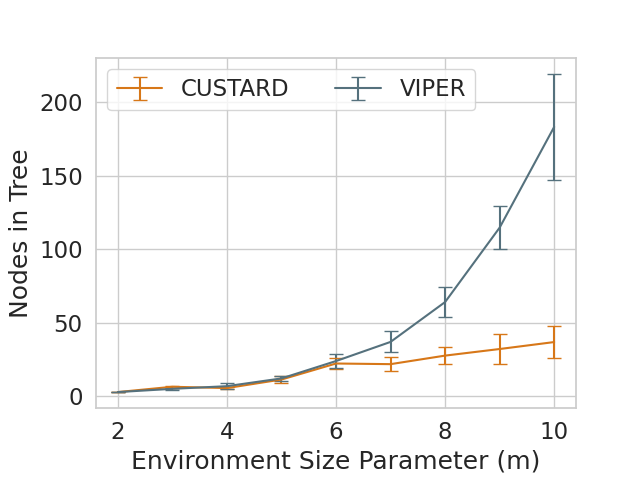}
         \label{fig:sizevsnodes}
     \end{subfigure}
    \caption{Tree depth and node count as the PrereqWorld environment size increases (Std Dev bars). \mname{} yields smaller trees for larger environments than VIPER.}
    \label{fig:res}
\end{figure}
    To evaluate \mname{}'s ability to produce DTPs with a non-interpretable function approximator for the IBMDP, we apply the \mname{} modifications to three base methods: 
    DDQN, PPO, 
    and MFEC 
    with improvements from Neural Episodic Control~\cite{pritzel2017neural}.
    DDQN is a Q-learning approach that uses a neural network to learn a state-value function and action-value function, which are combined to form a Q-function.
    PPO is a policy gradient method that uses a critic for estimating the advantage function. We use a neural network for both the actor and critic.
    MFEC is a Q-learning approach that uses a nearest neighbor model to estimate Q-values.
    
    The modifications from Section~\ref{approach_training_procedure} are applied to all three methods. 
    Actions are selected based on IBMDP states without the base state ($s_w \setminus s_b$); this affects the actor for PPO and the Q-function for DDQN and MFEC.
    DDQN and MFEC are used with a target Q-value function, $Q_o$, when performing updates, as in Figure~\ref{fig:omnqfn}.
    The target function and the critic for PPO are used with full IBMDP states.
    
    We compare to VIPER using two expert types: DQN and Backward Induction (BI).
    In Table~\ref{table:mname_table}, we show the final average reward and tree depth for 50 trials on CartPole, PrereqWorld ($m=7$), and PotholeWorld.
    Optimal final average rewards would be 200, -4, and 50, respectively. 
    \mname{} finds DTPs with high average reward for all environments and tends to find shorter DTPs than VIPER.
    To further evaluate depth-vs.-reward trade-offs, we use VIPER(BI) and \mname{}(MFEC) since these methods have the fewest hyperparameters and are least computationally expensive. 

\subsection{Response to Environment Size}
\label{sec:env_size}
    \mname{} discourages the learning of unnecessarily large trees through the use of two penalty terms, $\zeta$ and $\gamma_w$.
    These penalties are akin to regularization of the implicit DTP: when multiple optimal DTPs exist for the base MDP, the optimal IBMDP policy corresponds to the DTP with the lowest average leaf height.
    In contrast, if a tree mimics an expert policy, then the resulting tree will include complex behaviors that are artifacts of the expert's intricacy.
    
    We evaluate the decrease in tree size attained by using \mname{} to directly learn a tree.
    We compare the tree depth and node count for DTPs found by VIPER and \mname{} on PrereqWorld.
    The environment size is varied through $m$, which specifies the number of states ($2^m$) and the number of actions ($m$).
    For a given $m$, average reward is equal for both methods.
    The results are shown in Figure~\ref{fig:res} (50 trials per method/$m$ pair). 
    \mname{} produces smaller trees for $m\geq4$, and the size differences increases with the environment size.
    This is because an unconstrained expert can learn more complex behaviors with a larger state space, and VIPER faithfully mimics the expert policy.

\subsection{Response to Tree Depth}
\label{sec:tree_depth}
    \begin{figure}[t]
    \centering
         \includegraphics[width=0.95\columnwidth]{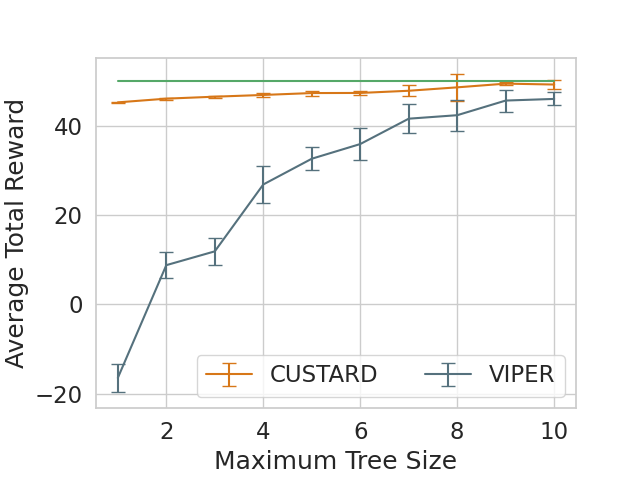}
    \caption{Average per-episode reward for trees of a fixed depth for PotholeWorld (Std Dev bars). \mname{} DTPs consistently achieve higher reward than VIPER's DTPs. Bar at 50 indicates maximum possible per-episode reward.}
    \label{fig:resdepth}
\end{figure}
    If an application requires a DTP of fixed depth, then fitting a DT to an expert policy can yield a poor policy of that depth.
    This is because the expert is not learned in the context of the depth limitation; imperfectly imitating that expert can lead to low reward.
    \mname{} yields better policies at a given depth since it directly solves an IBMDP that can be augmented with a depth limit.
    An IBMDP can include affordances~\cite{khetarpal2020can}, so that information-gathering actions cannot be chosen 
    $n$ actions after the most recent base action.
    With this modification, an RL algorithm can directly find the best DTP subject to the depth restriction.
    
    We evaluate \mname{}'s ability to find DTPs with high average reward for PotholeWorld subject to a tree depth limit.
    This domain is designed so the overall optimal DTP cannot be pruned to obtain the optimal DTP for a smaller depth.
    We present the average episode reward as a function of the depth limit in Figure~\ref{fig:resdepth} for VIPER and \mname{} (50 trials per method/depth pair).
    \mname{} attains higher reward through using $lane\_1$ when the DTP depth is too shallow to avoid potholes in the other lanes.
    In contrast, VIPER always attempts to imitate the expert and attains a low reward when the DTP poorly represents the expert policy.

\section{Conclusion and Future Work}

We introduce Iterative Bounding MDPs, an MDP representation which corresponds to the problem of finding a decision tree policy for an underlying MDP.
Additionally, we identify how the standard value update rule must be changed so that all IBMDP solutions correspond to decision tree policies for the underlying MDP.
We show how existing RL algorithms can be modified to solve IBMDPs, so a non-interpretable function approximator can be used in conjunction with an existing RL method to solve an IBMDP and produce a decision tree policy.
In addition, we provide empirical results showing the tree size and reward improvements possible through solving an IBMDP rather than approximating a non-interpretable expert.
Future work includes generalization of IBMDPs to encode other constraints 
so other explanation formats can be directly learned.

\section{Acknowledgements}
This material is based upon work supported by the Department of Defense (DoD) through the National Defense Science \& Engineering Graduate (NDSEG) Fellowship Program, DARPA/AFRL agreement FA87501720152, and NSF grant IIS-1850477. Any opinions, findings and conclusions or recommendations expressed in this material are those of the author(s) and do not necessarily reflect the views of the Department of Defense, Defense Advanced Research Projects Agency, the Air Force Research Laboratory, or the National Science Foundation.

\bibliography{aaai21-shorter}

\begin{thebibliography}{62}
\providecommand{\natexlab}[1]{#1}
\providecommand{\url}[1]{\texttt{#1}}
\providecommand{\urlprefix}{URL }
\expandafter\ifx\csname urlstyle\endcsname\relax
  \providecommand{\doi}[1]{doi:\discretionary{}{}{}#1}\else
  \providecommand{\doi}{doi:\discretionary{}{}{}\begingroup
  \urlstyle{rm}\Url}\fi

\bibitem[{Abadi et~al.(2015)Abadi, Agarwal, Barham, Brevdo, Chen, Citro,
  Corrado, Davis, Dean, Devin, Ghemawat, Goodfellow, Harp, Irving, Isard, Jia,
  Jozefowicz, Kaiser, Kudlur, Levenberg, Man\'{e}, Monga, Moore, Murray, Olah,
  Schuster, Shlens, Steiner, Sutskever, Talwar, Tucker, Vanhoucke, Vasudevan,
  Vi\'{e}gas, Vinyals, Warden, Wattenberg, Wicke, Yu, and
  Zheng}]{tensorflow2015-whitepaper}
Abadi, M.; Agarwal, A.; Barham, P.; Brevdo, E.; Chen, Z.; Citro, C.; Corrado,
  G.~S.; Davis, A.; Dean, J.; Devin, M.; Ghemawat, S.; Goodfellow, I.; Harp,
  A.; Irving, G.; Isard, M.; Jia, Y.; Jozefowicz, R.; Kaiser, L.; Kudlur, M.;
  Levenberg, J.; Man\'{e}, D.; Monga, R.; Moore, S.; Murray, D.; Olah, C.;
  Schuster, M.; Shlens, J.; Steiner, B.; Sutskever, I.; Talwar, K.; Tucker, P.;
  Vanhoucke, V.; Vasudevan, V.; Vi\'{e}gas, F.; Vinyals, O.; Warden, P.;
  Wattenberg, M.; Wicke, M.; Yu, Y.; and Zheng, X. 2015.
\newblock {TensorFlow}: Large-Scale Machine Learning on Heterogeneous Systems.
\newblock \urlprefix\url{http://tensorflow.org/}.
\newblock Software available from tensorflow.org.

\bibitem[{Anderson et~al.(2019)Anderson, Dodge, Sadarangani, Juozapaitis,
  Newman, Irvine, Chattopadhyay, Fern, and Burnett}]{anderson2019explaining}
Anderson, A.; Dodge, J.; Sadarangani, A.; Juozapaitis, Z.; Newman, E.; Irvine,
  J.; Chattopadhyay, S.; Fern, A.; and Burnett, M. 2019.
\newblock Explaining reinforcement learning to mere mortals: An empirical
  study.
\newblock In \emph{Proc. of the 28th Int. Joint Conf. on Artificial
  Intelligence}.

\bibitem[{Annasamy and Sycara(2019)}]{annasamy2019towards}
Annasamy, R.~M.; and Sycara, K. 2019.
\newblock Towards better interpretability in deep q-networks.
\newblock In \emph{Proc. of the 33rd AAAI Conf. on Artificial Intelligence}.

\bibitem[{Atrey, Clary, and Jensen(2020)}]{atrey2019exploratory}
Atrey, A.; Clary, K.; and Jensen, D. 2020.
\newblock Exploratory Not Explanatory: Counterfactual Analysis of Saliency Maps
  for Deep RL.
\newblock In \emph{Proc. of the 8th Int. Conf. on Learning Representations}.

\bibitem[{Barto, Sutton, and Anderson(1983)}]{barto1983neuronlike}
Barto, A.~G.; Sutton, R.~S.; and Anderson, C.~W. 1983.
\newblock Neuronlike adaptive elements that can solve difficult learning
  control problems.
\newblock \emph{IEEE Transactions on Systems, Man, and Cybernetics} (5).

\bibitem[{Bastani, Pu, and Solar{-}Lezama(2018)}]{bastani2018viper}
Bastani, O.; Pu, Y.; and Solar{-}Lezama, A. 2018.
\newblock Verifiable Reinforcement Learning via Policy Extraction.
\newblock In \emph{Advances in Neural Information Processing Systems}.

\bibitem[{Blundell et~al.(2016)Blundell, Uria, Pritzel, Li, Ruderman, Leibo,
  Rae, Wierstra, and Hassabis}]{blundell2016model}
Blundell, C.; Uria, B.; Pritzel, A.; Li, Y.; Ruderman, A.; Leibo, J.~Z.; Rae,
  J.; Wierstra, D.; and Hassabis, D. 2016.
\newblock Model-free episodic control.
\newblock \emph{arXiv preprint arXiv:1606.04460} .

\bibitem[{Boutilier et~al.(1995)Boutilier, Dearden, Goldszmidt
  et~al.}]{boutilier1995exploiting}
Boutilier, C.; Dearden, R.; Goldszmidt, M.; et~al. 1995.
\newblock Exploiting structure in policy construction.
\newblock In \emph{Proc. of the 14th Int. Joint Conf. on Artificial
  Intelligence}.

\bibitem[{Bradtke and Duff(1994)}]{bradtke1994reinforcement}
Bradtke, S.; and Duff, M. 1994.
\newblock Reinforcement learning methods for continuous-time {M}arkov decision
  problems.
\newblock \emph{Advances in Neural Information Processing Systems} .

\bibitem[{Brockman et~al.(2016)Brockman, Cheung, Pettersson, Schneider,
  Schulman, Tang, and Zaremba}]{brockman2016gym}
Brockman, G.; Cheung, V.; Pettersson, L.; Schneider, J.; Schulman, J.; Tang,
  J.; and Zaremba, W. 2016.
\newblock OpenAI Gym.
\newblock \emph{arXiv preprint arXiv:1606.01540} .

\bibitem[{Buciluǎ, Caruana, and Niculescu-Mizil(2006)}]{bucilua2006model}
Buciluǎ, C.; Caruana, R.; and Niculescu-Mizil, A. 2006.
\newblock Model compression.
\newblock In \emph{Proc. of the 12th ACM Int. Conf. on Knowledge Discovery and
  Data Mining}.

\bibitem[{Cruz, Dazeley, and Vamplew(2019)}]{cruz2019memorybasedxrl}
Cruz, F.; Dazeley, R.; and Vamplew, P. 2019.
\newblock Memory-Based Explainable Reinforcement Learning.
\newblock \emph{AI 2019: Advances in Artificial Intelligence} .

\bibitem[{Dayan and Hinton(1993)}]{dayan1993feudal}
Dayan, P.; and Hinton, G.~E. 1993.
\newblock Feudal reinforcement learning.
\newblock In \emph{Advances in Neural Information Processing Systems}.

\bibitem[{Degris, Sigaud, and Wuillemin(2006)}]{degris2006learning}
Degris, T.; Sigaud, O.; and Wuillemin, P.-H. 2006.
\newblock Learning the structure of factored {M}arkov decision processes in
  reinforcement learning problems.
\newblock In \emph{Proc. of the 23rd Int. Conf. on Machine Learning}.

\bibitem[{Dietterich(2000)}]{dietterich2000maxq}
Dietterich, T.~G. 2000.
\newblock Hierarchical Reinforcement Learning with the MAXQ Value Function
  Decomposition.
\newblock \emph{Journal of Artificial Intelligence Research} 13.

\bibitem[{Dodson, Mattei, and Goldsmith(2011)}]{dodson2011natural}
Dodson, T.; Mattei, N.; and Goldsmith, J. 2011.
\newblock A natural language argumentation interface for explanation generation
  in {M}arkov decision processes.
\newblock In \emph{Proc. of the 2nd Int. Conf. on Algorithmic Decision Theory}.

\bibitem[{Greydanus et~al.(2018)Greydanus, Koul, Dodge, and
  Fern}]{greydanus2018visualizing}
Greydanus, S.; Koul, A.; Dodge, J.; and Fern, A. 2018.
\newblock Visualizing and Understanding Atari Agents.
\newblock In \emph{Proc. of the 35th Int. Conf. on Machine Learning}.

\bibitem[{Hein et~al.(2017)Hein, Hentschel, Runkler, and
  Udluft}]{HEIN2017fuzzyrl}
Hein, D.; Hentschel, A.; Runkler, T.; and Udluft, S. 2017.
\newblock Particle swarm optimization for generating interpretable fuzzy
  reinforcement learning policies.
\newblock \emph{Eng. Applications of Artificial Intelligence} 65.

\bibitem[{Hein, Udluft, and Runkler(2019)}]{genetic}
Hein, D.; Udluft, S.; and Runkler, T.~A. 2019.
\newblock Interpretable Policies for Reinforcement Learning by Genetic
  Programming.
\newblock In \emph{Proc. of the Genetic and Evolutionary Computation Conf.}

\bibitem[{Hester, Quinlan, and Stone(2010)}]{hester2010generalized}
Hester, T.; Quinlan, M.; and Stone, P. 2010.
\newblock Generalized model learning for reinforcement learning on a humanoid
  robot.
\newblock In \emph{Proc. of the Int. Conf. on Robotics and Automation}.

\bibitem[{Hill et~al.(2018)Hill, Raffin, Ernestus, Gleave, Kanervisto, Traore,
  Dhariwal, Hesse, Klimov, Nichol, Plappert, Radford, Schulman, Sidor, and
  Wu}]{stable-baselines}
Hill, A.; Raffin, A.; Ernestus, M.; Gleave, A.; Kanervisto, A.; Traore, R.;
  Dhariwal, P.; Hesse, C.; Klimov, O.; Nichol, A.; Plappert, M.; Radford, A.;
  Schulman, J.; Sidor, S.; and Wu, Y. 2018.
\newblock Stable Baselines.
\newblock \url{https://github.com/hill-a/stable-baselines}.

\bibitem[{Hinton, Vinyals, and Dean(2015)}]{hinton2015distilling}
Hinton, G.; Vinyals, O.; and Dean, J. 2015.
\newblock Distilling the knowledge in a neural network.
\newblock \emph{arXiv preprint arXiv:1503.02531} .

\bibitem[{Huber, Schiller, and Andr{\'e}(2019)}]{huber2019enhancing}
Huber, T.; Schiller, D.; and Andr{\'e}, E. 2019.
\newblock Enhancing explainability of deep reinforcement learning through
  selective layer-wise relevance propagation.
\newblock In \emph{Proc. of the Joint German/Austrian Conf. on Artificial
  Intelligence (K{\"u}nstliche Intelligenz)}.

\bibitem[{Iyer et~al.(2018)Iyer, Li, Li, Lewis, Sundar, and Sycara}]{katia}
Iyer, R.; Li, Y.; Li, H.; Lewis, M.; Sundar, R.; and Sycara, K. 2018.
\newblock Transparency and Explanation in Deep Reinforcement Learning Neural
  Networks.
\newblock In \emph{Proc. of the 1st AAAI/ACM Conf. on Artificial Intelligence,
  Ethics, and Society}.

\bibitem[{Juozapaitis et~al.(2019)Juozapaitis, Koul, Fern, Erwig, and
  Doshi-Velez}]{juozapaitis2019explainable}
Juozapaitis, Z.; Koul, A.; Fern, A.; Erwig, M.; and Doshi-Velez, F. 2019.
\newblock Explainable reinforcement learning via reward decomposition.
\newblock In \emph{28th Int. Joint Conf. on Artificial Intelligence Workshop on
  Explainable Artificial Intelligence}.

\bibitem[{Khetarpal et~al.(2020)Khetarpal, Ahmed, Comanici, Abel, and
  Precup}]{khetarpal2020can}
Khetarpal, K.; Ahmed, Z.; Comanici, G.; Abel, D.; and Precup, D. 2020.
\newblock What can I do here? A Theory of Affordances in Reinforcement
  Learning.
\newblock \emph{arXiv preprint arXiv:2006.15085} .

\bibitem[{Koul, Greydanus, and Fern(2018)}]{koul2018learning}
Koul, A.; Greydanus, S.; and Fern, A. 2018.
\newblock Learning finite state representations of recurrent policy networks.
\newblock \emph{arXiv preprint, arXiv:1811.12530} .

\bibitem[{Lipton(2018)}]{lipton2018mythos}
Lipton, Z.~C. 2018.
\newblock The mythos of model interpretability.
\newblock \emph{ACM Queue} 16(3).

\bibitem[{Madumal et~al.(2020{\natexlab{a}})Madumal, Miller, Sonenberg, and
  Vetere}]{madumal2020distal}
Madumal, P.; Miller, T.; Sonenberg, L.; and Vetere, F. 2020{\natexlab{a}}.
\newblock Distal Explanations for Explainable Reinforcement Learning Agents.
\newblock \emph{arXiv preprint arXiv:2001.10284} .

\bibitem[{Madumal et~al.(2020{\natexlab{b}})Madumal, Miller, Sonenberg, and
  Vetere}]{madumal2020explainable}
Madumal, P.; Miller, T.; Sonenberg, L.; and Vetere, F. 2020{\natexlab{b}}.
\newblock Explainable reinforcement learning through a causal lens.
\newblock In \emph{Proc. of the 34th AAAI Conf. on Artificial Intelligence}.

\bibitem[{Mahadevan et~al.(1997)Mahadevan, Marchalleck, Das, and
  Gosavi}]{mahadevan1997self}
Mahadevan, S.; Marchalleck, N.; Das, T.~K.; and Gosavi, A. 1997.
\newblock Self-improving factory simulation using continuous-time
  average-reward reinforcement learning.
\newblock In \emph{Proc. of the 14th Int. Conf. on Machine Learning}.

\bibitem[{Masson, Ranchod, and Konidaris(2016)}]{masson2016paramactionrl}
Masson, W.; Ranchod, P.; and Konidaris, G. 2016.
\newblock Reinforcement Learning with Parameterized Actions.
\newblock In \emph{Proc. of the 30th AAAI Conf. on Artificial Intelligence}.

\bibitem[{McCallum(1997)}]{mccallum1997reinforcement}
McCallum, R. 1997.
\newblock Reinforcement learning with selective perception and hidden state.
\newblock \emph{PhD Thesis, University of Rochester, Department of Computer
  Science} .

\bibitem[{Milani et~al.(2020)Milani, Topin, Houghton, Guss, Mohanty, Nakata,
  Vinyals, and Kuno}]{milani2020minerl}
Milani, S.; Topin, N.; Houghton, B.; Guss, W.~H.; Mohanty, S.~P.; Nakata, K.;
  Vinyals, O.; and Kuno, N.~S. 2020.
\newblock Retrospective Analysis of the 2019 {MineRL} Competition on
  Sample-Efficient Reinforcement Learning Using Human Priors.
\newblock \emph{NeurIPS2019 Competition \& Demonstration Track Postproceedings}
  .

\bibitem[{Mnih et~al.(2013)Mnih, Kavukcuoglu, Silver, Graves, Antonoglou,
  Wierstra, and Riedmiller}]{mnih2013playing}
Mnih, V.; Kavukcuoglu, K.; Silver, D.; Graves, A.; Antonoglou, I.; Wierstra,
  D.; and Riedmiller, M. 2013.
\newblock Playing {A}tari with deep reinforcement learning.
\newblock \emph{arXiv preprint arXiv:1312.5602} .

\bibitem[{Molnar(2019)}]{molnar2020interpretable}
Molnar, C. 2019.
\newblock \emph{Interpretable Machine Learning}.
\newblock \url{https://christophm.github.io/interpretable-ml-book/}.

\bibitem[{Parr and Russell(1998)}]{parr1998hierarchical}
Parr, R.~E.; and Russell, S. 1998.
\newblock \emph{Hierarchical control and learning for Markov decision
  processes}.
\newblock University of California, Berkeley Berkeley, CA.

\bibitem[{Precup, Sutton, and Singh(1998)}]{precup1998theoretical}
Precup, D.; Sutton, R.~S.; and Singh, S. 1998.
\newblock Theoretical results on reinforcement learning with temporally
  abstract options.
\newblock In \emph{European Conf. on Machine Learning}.

\bibitem[{Preda(2007)}]{preda2007buildingdtbyrl}
Preda, M. 2007.
\newblock Adaptive Building of Decision Trees by Reinforcement Learning.
\newblock In \emph{Proc. of the 7th Int. Conf. on Applied Informatics and
  Communications}.

\bibitem[{Pritzel et~al.(2017)Pritzel, Uria, Srinivasan, Puigdomenech, Vinyals,
  Hassabis, Wierstra, and Blundell}]{pritzel2017neural}
Pritzel, A.; Uria, B.; Srinivasan, S.; Puigdomenech, A.; Vinyals, O.; Hassabis,
  D.; Wierstra, D.; and Blundell, C. 2017.
\newblock Neural episodic control.
\newblock \emph{arXiv preprint arXiv:1703.01988} .

\bibitem[{Pyeatt(2003)}]{pyeatt2003reinforcement}
Pyeatt, L.~D. 2003.
\newblock Reinforcement Learning with Decision Trees.
\newblock In \emph{Applied Informatics}.

\bibitem[{Pyeatt and Howe(2001)}]{pyeatt2001decision}
Pyeatt, L.~D.; and Howe, A.~E. 2001.
\newblock Decision tree function approximation in reinforcement learning.
\newblock In \emph{Proc. of the 3rd Int. Symposium on Adaptive Systems:
  Evolutionary Computation and Probabilistic Graphical Models}.

\bibitem[{Quinlan(1986)}]{quinlan1986induction}
Quinlan, J.~R. 1986.
\newblock Induction of decision trees.
\newblock \emph{Machine learning} 1(1).

\bibitem[{Rodriguez et~al.(2020)Rodriguez, Killian, Son, and
  Gombolay}]{rodriguez2020optimization}
Rodriguez, I. D.~J.; Killian, T.~W.; Son, S.; and Gombolay, M.~C. 2020.
\newblock Optimization Methods for Interpretable Differentiable Decision Trees
  in Reinforcement Learning.
\newblock In \emph{Proc. of the 23rd Int. Conf. on Artificial Intelligence and
  Statistics}.

\bibitem[{Ross, Gordon, and Bagnell(2011)}]{ross2011dagger}
Ross, S.; Gordon, G.~J.; and Bagnell, J.~A. 2011.
\newblock A Reduction of Imitation Learning and Structured Prediction to
  No-Regret Online Learning.
\newblock In \emph{Proc. of the 14th Int. Conf. on Artificial Intelligence and
  Statistics}.

\bibitem[{Roth et~al.(2019)Roth, Topin, Jamshidi, and
  Veloso}]{roth2019conservative}
Roth, A.~M.; Topin, N.; Jamshidi, P.; and Veloso, M. 2019.
\newblock Conservative Q-Improvement: Reinforcement Learning for an
  Interpretable Decision-Tree Policy.
\newblock \emph{arXiv preprint arXiv:1907.01180} .

\bibitem[{Rusu et~al.(2015)Rusu, Colmenarejo, Gulcehre, Desjardins,
  Kirkpatrick, Pascanu, Mnih, Kavukcuoglu, and Hadsell}]{rusu2015policy}
Rusu, A.~A.; Colmenarejo, S.~G.; Gulcehre, C.; Desjardins, G.; Kirkpatrick, J.;
  Pascanu, R.; Mnih, V.; Kavukcuoglu, K.; and Hadsell, R. 2015.
\newblock Policy distillation.
\newblock \emph{arXiv preprint arXiv:1511.06295} .

\bibitem[{Schulman et~al.(2017)Schulman, Wolski, Dhariwal, Radford, and
  Klimov}]{schulman2017proximal}
Schulman, J.; Wolski, F.; Dhariwal, P.; Radford, A.; and Klimov, O. 2017.
\newblock Proximal policy optimization algorithms.
\newblock \emph{arXiv preprint arXiv:1707.06347} .

\bibitem[{Sokol and Flach(2019)}]{sokol2019desiderata}
Sokol, K.; and Flach, P. 2019.
\newblock Desiderata for Interpretability: Explaining Decision Tree Predictions
  with Counterfactuals.
\newblock In \emph{Proc. of the 33rd AAAI Conf. on Artificial Intelligence,
  Student Abstract Track}.

\bibitem[{Sreedharan, Srivastava, and Kambhampati(2020)}]{sreedharan2020tldr}
Sreedharan, S.; Srivastava, S.; and Kambhampati, S. 2020.
\newblock TLdR: Policy Summarization for Factored SSP Problems Using Temporal
  Abstractions.
\newblock In \emph{Proc. of the 30th Int. Conf. on Automated Planning and
  Scheduling}.

\bibitem[{Strehl, Diuk, and Littman(2007)}]{strehl2007efficient}
Strehl, A.~L.; Diuk, C.; and Littman, M.~L. 2007.
\newblock Efficient structure learning in factored-state MDPs.
\newblock In \emph{Proc. of the 22nd AAAI Conf. on Artificial Intelligence}.

\bibitem[{Sutton, Precup, and Singh(1999)}]{sutton1999between}
Sutton, R.~S.; Precup, D.; and Singh, S. 1999.
\newblock Between MDPs and semi-MDPs: A framework for temporal abstraction in
  reinforcement learning.
\newblock \emph{Artificial Intelligence} 112(1-2).

\bibitem[{Tabrez, Agrawal, and Hayes(2019)}]{tabrez2019explanation}
Tabrez, A.; Agrawal, S.; and Hayes, B. 2019.
\newblock Explanation-based reward coaching to improve human performance via
  reinforcement learning.
\newblock In \emph{Proc. of the 14th Int. Conf. on Human-Robot Interaction}.

\bibitem[{Tang, Nguyen, and Ha(2020)}]{tang2020neuroevolution}
Tang, Y.; Nguyen, D.; and Ha, D. 2020.
\newblock Neuroevolution of Self-Interpretable Agents.
\newblock \emph{arXiv preprint arXiv:2003.08165} .

\bibitem[{Tieleman and Hinton(2012)}]{tieleman2012lecture}
Tieleman, T.; and Hinton, G. 2012.
\newblock Lecture 6.5-rmsprop: Divide the gradient by a running average of its
  recent magnitude.
\newblock \emph{COURSERA: Neural networks for machine learning} 4(2).

\bibitem[{Topin and Veloso(2019)}]{topin2019xdrl}
Topin, N.; and Veloso, M. 2019.
\newblock Generation of Policy-Level Explanations for Reinforcement Learning.
\newblock In \emph{Proc. of the 33rd AAAI Conf. on Artificial Intelligence}.

\bibitem[{Tuyls, Maes, and Manderick(2002)}]{tuyls2002reinforcement}
Tuyls, K.; Maes, S.; and Manderick, B. 2002.
\newblock Reinforcement learning in large state spaces.
\newblock In \emph{Robot Soccer World Cup}.

\bibitem[{Uther and Veloso(1998)}]{uther1998tree}
Uther, W.~T.; and Veloso, M.~M. 1998.
\newblock Tree based discretization for continuous state space reinforcement
  learning.
\newblock In \emph{Proc. of the 15th National Conf. on Artificial
  Intelligence}.

\bibitem[{Vasic et~al.(2019)Vasic, Petrovic, Wang, Nikolic, Singh, and
  Khurshid}]{vasic2019moet}
Vasic, M.; Petrovic, A.; Wang, K.; Nikolic, M.; Singh, R.; and Khurshid, S.
  2019.
\newblock Mo{\"{E}}T: Interpretable and Verifiable Reinforcement Learning via
  Mixture of Expert Trees.
\newblock \emph{arXiv preprint arXiv:1906.06717} .

\bibitem[{Wang et~al.(2018)Wang, Chen, Yang, Wu, Wu, and
  Xie}]{wang2018reinforcement}
Wang, X.; Chen, Y.; Yang, J.; Wu, L.; Wu, Z.; and Xie, X. 2018.
\newblock A reinforcement learning framework for explainable recommendation.
\newblock In \emph{Proc. of the Int. Conf. on Data Mining}.

\bibitem[{Wang et~al.(2016)Wang, Schaul, Hessel, Hasselt, Lanctot, and
  Freitas}]{wang2016dueling}
Wang, Z.; Schaul, T.; Hessel, M.; Hasselt, H.; Lanctot, M.; and Freitas, N.
  2016.
\newblock Dueling network architectures for deep reinforcement learning.
\newblock In \emph{Proc. of the 33rd Int. Conf. on Machine Learning}.

\bibitem[{Yang et~al.(2018)Yang, Bai, Zhang, and Torr}]{yang2018learn}
Yang, Z.; Bai, S.; Zhang, L.; and Torr, P.~H. 2018.
\newblock Learn to interpret {A}tari agents.
\newblock \emph{arXiv preprint arXiv:1812.11276} .

\end{thebibliography}
\newpage
\appendix
\appendixpage
\section{Environments}
\label{sec:env-desc}
\subsection{CartPole}
The CartPole~\cite{brockman2016gym} domain, first described by~\citet{barto1983neuronlike}, consists of a pole affixed to a cart that can move along a single dimension. 
The agent must balance the pole in an upright position by moving the cart back and forth along a frictionless track.
The state features are the cart position, the cart velocity, the pole angle, and the pole velocity at tip.
The cart position is bounded by $[-2.4, 2.4]$, the cart velocity is bounded by $[\infty, \infty] $, the pole angle is bounded by $[-41.8, 41.8]$, and the pole velocity at tip is bounded by $[-\infty, \infty ]$.

To create the start state, each feature is assigned a uniform random value between $\pm 0.05$.
The two actions are push the cart to the left or push the cart to the right.
The reward is $+1$ for each timestep where the pole remains upright.
An episode terminates when the pole falls (is greater than $15$ degrees from vertical), when $200$ timesteps elapse, or when the cart reaches the end of the display (the cart position is more than $\pm 2.4 $).

\subsection{PrereqWorld}
The PrereqWorld~\cite{topin2019xdrl} domain is an abstraction of a production task in which the agent is tasked with creating a goal item.
Constructing each item may require a subset of other, prerequisite items, and the agent may only have one of each item type at a time.
    The items are topologically sorted based on their prerequisite relationships such that lower-numbered items require higher-numbered items.
    A PrereqWorld environment has a number of item types $m$
     and a stochasticity parameter $\rho$, which we set to zero for a deterministic environment.
    A state consists of $m$ binary features which indicate whether each type of item is present.
    There are $m$ actions: each action corresponds to an attempt to produce a specific item.
    If the required items are present, then they are consumed (features set to $0$) and the new item is created (feature set to 1).
    If not all required items are present, the action has no effect (i.e., the next state is the same as the current state).
    Creating the goal item yields a reward of $0$ and ends the episode; all other actions yield a reward of -1.

    We use a base PrereqWorld environment with $m=10$ with a fixed set of prerequisites.
    The prerequisites are (in `source: [prerequisite list]' format): 0: [1, 2,], 1: [5], 2: [4], 3: [6, 7], 4: [], 5: [7], 6: [7, 9], 7: [8], 8: [], 9: [].
    We produce smaller variants by removing high-numbed items (based on the topological sort). 
    
\subsection{PotholeWorld}
We introduce the PotholeWorld domain, in which
the agent is tasked with traversing $50$ units along a three-lane road.
The first lane gives less reward per unit traveled; however, the other two lanes contain ``potholes'' that lead to a reward penalty if traversed.
A state contains a single feature: the current position, in $[0,50]$. 
The initial state is at position $0$; an episode terminates when the position is $50$.
The three actions are $[lane\_1, lane\_2, lane\_3]$, which each advance the agent a random amount drawn from $Unif(0.5,1)$.

Potholes are added starting from position $0$ until position $50$ with distances drawn from $Unif(1,2)$. 
Potholes are assigned with equal probability to lane two or lane three. 
The agent cannot directly observe the potholes; their locations must be learned through repeated interactions with the environment.
When an action is taken, the base reward is equal to the distance moved. This is reduced by $10\%$ if the $lane\_1$ action was taken.
If a pothole is in the chosen lane along the traversed stretch (i.e., between previous agent position and next agent position), then the reward is reduced by $5$.

For our evaluation, our generated PotholeWorld domain contains potholes in the following locations.
The locations for potholes in lane two are: [3.09, 5.97, 7.33, 8.874, 9.98, 11.70, 12.83, 14.62, 16.33, 19.55, 27.56, 31.28, 33.07, 36.30, 37.81, 39.14, 44.21, 46.81, 49.05].
The locations for potholes in lane three are: [0, 1.30, 4.53, 17.45, 18.47, 21.42, 23.34, 24.42, 25.70, 29.41, 34.46, 40.45, 42.39, 45.30, 47.87].

\section{Experimental Settings}
\label{sec:exp-settings}

\subsection{VIPER (DQN)}
We use DQN~\cite{mnih2013playing} as the expert for VIPER (DQN).
For the VIPER (DQN) experiments, we use the standard parameters from the Stable Baselines~\cite{stable-baselines} implementation of DQN.
This implementation of DQN, by default, uses a multi-layer perceptron with two hidden layers, each with $64$ units. 
For training DQN, we use $10^5$, $10^5$, and $10^6$ training steps for CartPole, PrereqWorld, and PotholeWorld respectively. 

For the VIPER~\cite{bastani2018viper} tree extraction, we follow the parameter guidelines from the original paper and use: $10$ batch rollouts, $200,000$ maximum samples, $300$ maximum iterations, a training fraction of $0.8$, and $50$ test rollouts.
These values are chosen since they produced high-performing tree policies, and higher values produce similar policies. 

\subsection{VIPER (BI)}
We use the standard Backward Induction (BI) algorithm as the expert for VIPER (BI).
For VIPER (BI) on PotholeWorld, we break ties by favoring the previously chosen (downstream) action.
This is done in PotholeWorld to avoid oscillation within the expert; this produces simpler-form policies which still attain optimal reward.
For VIPER (BI) on PrereqWorld, we break ties at random since this oscillation behavior cannot arise in this domain (i.e., the same action is never repeated twice in sequence).

\subsection{RLDTRL (DQN)}
For RLDTRL (DQN) we use a custom Tensorflow~\cite{tensorflow2015-whitepaper} DDQN~\cite{wang2016dueling} implementation as a starting point.
We use memory replay, with a buffer size of $10^6$, filled with $10^5$ random policy transitions before training.
We use $\epsilon$-greedy exploration, and decay $\epsilon$ linearly from $0.5$ to $0.05$ over the course of $2 \times 10^5$ transitions.
We use the RMSProp~\cite{tieleman2012lecture} optimizer with a learning rate of $2.5 \times 10^{-4}$ and a $\rho$ of 0.95. 
We train for $10^6$ episodes, with batches of size 128, using two hidden layers of width 128.

For CartPole, we use a $p$ of 3, a $\zeta$ of -0.01, $\gamma_b$ and $\gamma_w$ of 1, and, for the purposes of the IBMDP, treat the continuous features as upper-bounded by $[2,2,0.14,1.4]$ and lower-bounded by $[-2,-2,-0.14, -1.4]$.
For PrereqWorld, we use a $p$ of 1, a $\zeta$ of -0.01, and $\gamma_b$ and $\gamma_w$ of 1. 
For PotholeWorld, we use a $p$ of 3, a $\zeta$ of 0, and $\gamma_b$ and $\gamma_w$ of 0.99.

\subsection{RLDTRL (PPO)}
We use the PPO implementation from Stable Baselines~\cite{stable-baselines} as a starting point.
Unless otherwise specified, the hyperparameters are the default ones specified in the Stable Baselines implementation.
For all environments, we use two hidden layers of $64$ units each for both the actor and the critic (no shared layers). 
We use $4$ minibatches for all environments and train for a maximum of $3.2 \times 10^6$ timesteps.

For CartPole, we use a $p$ of 3. For the purposes of the IBMDP, we treat CartPole's features as bounded above by $[2,2,0.14,1.4]$ and below by $[-2,-2,-0.14,-1.4]$. 
We tried a looser bound ($10$ above and $-10$ below for all features), but the resulting trees were deeper. 
We use $512$ steps per batch. 
We use $\gamma_w$ and $\gamma_b$ of 1, and we tried setting both to $0.999$.
We use a $\zeta$ of -0.01.

For PrereqWorld, we use a $p$ of 1.
For the presented results, we use $\gamma_w$ and $\gamma_b$ of $1$. We also tried values of $0.99$ but had faster training with $1$.
We use a $\zeta$ of $-0.01$.
For the neural network architecture, we also tried three hidden layers of $64$ and two hidden layers of $128$ units.
We use $1024$ steps per batch.

For PotholeWorld, we use a $p$ of 3. We also tried a $p$ of 10, but this caused the agent to have difficulty learning a good policy. We use $\gamma_w$ and $\gamma_b$ of 0.99, and we also tried $0.999$ for both. We use a $\zeta$ of 0, and we also tried $-0.01$. With the lower $\zeta$, the optimal policy is found less frequently. We also tried three hidden layers of $64$ and two hidden layers of $128$ units. 
We tried a learning rate of $0.001$.
We use $1024$ steps per batch.

\subsection{RLDTRL (MFEC)}
We replace the MFEC update with an NEC-style update~\cite{pritzel2017neural}.
MFEC's update uses the maximum of the previous value and the new estimate, which is suitable for deterministic environments. 
Given the uncertainty introduced by the IBMDP representation, we use a typical Q-function update using a learning rate.
We choose different learning rates for the policy Q-function ($\alpha_b$) and for the omniscient Q-function ($\alpha_o$), with a higher learning rate for the omniscient Q-function given that it is used to update the policy Q-function. To determine the learning rates, we tried $1$, $0.7$, $0.3$, $0.1$, $0.01$, and $0.001$ in decreasing order (with $\alpha_o$ two steps larger than $\alpha_b$) until a configuration was found that converged.
For all domains, we use $\zeta$ of $-0.01$ and $\gamma_w$ of $1$. 

For CartPole, we set $\gamma_b$ to $0.99$ and $k$ to $9$. We use a $p$ of 10.
We set $\alpha_b = 0.01$ and $ \alpha_o = 0.3$.  We trained for $10^6$ episodes.
For PrereqWorld, we set $\gamma_b$ to $1$ and $k$ to $9$. We use a $p$ of 1.
We set $\alpha_b = 0.1$ and $ \alpha_o = 0.7$. We trained for $2000 \times 2^m$ episodes.
For PotholeWorld, we set $\gamma_b$ to $1$ and $k$ to $2$.
We use a $p$ of 10.
We set $\alpha_b = 0.01$ and $ \alpha_o = 0.3$.
We trained for $10^6$ episodes.

\end{document}